\documentclass[%
 aip,
 amsmath,amssymb,
 reprint,%
]{revtex4-1}

\usepackage{graphicx}
\usepackage{dcolumn}
\usepackage{bm}

\usepackage[utf8]{inputenc}
\usepackage[T1]{fontenc}
\usepackage{mathptmx}
\usepackage{etoolbox}
\usepackage{xurl}
\usepackage{hyperref}
\usepackage{cleveref}
\usepackage{tikz}
\usetikzlibrary{calc,matrix,positioning,arrows,decorations.pathreplacing,angles,quotes,math}
\usepackage{booktabs}
\usepackage{dirtytalk} 
\usepackage{csquotes}
\usepackage{subfig}
\usepackage{tabularx}
\usepackage{multirow}
\usepackage{dirtytalk} 
\usepackage{placeins}
\usepackage{xcolor}
\definecolor{blue_source}{rgb}{0.4, 0.55, 0.7}
\definecolor{orange_sink}{rgb}{0.925,0.447,0.2}
\definecolor{green_stable}{HTML}{83c855}
\definecolor{purple_trouble}{HTML}{aa3377}
\definecolor{blue_perturbation_space}{HTML}{5cceff}

\makeatletter
\def\@email#1#2{%
 \endgroup
 \patchcmd{\titleblock@produce}
  {\frontmatter@RRAPformat}
  {\frontmatter@RRAPformat{\produce@RRAP{*#1\href{mailto:#2}{#2}}}\frontmatter@RRAPformat}
  {}{}
}%
\makeatother
\begin{document}

\preprint{AIP/123-QED}

\title[Toward Dynamic Stability Assessment of Power Grid Topologies using Graph Neural Networks]{Toward Dynamic Stability Assessment of Power Grid Topologies using Graph Neural Networks}

\author{Christian Nauck}
\affiliation{%
Potsdam Institute for Climate Impact Research, Telegrafenberg A31, 14473 Potsdam, Germany
}%

\author{Michael Lindner}%
\affiliation{%
Potsdam Institute for Climate Impact Research, Telegrafenberg A31, 14473 Potsdam, Germany
}

\author{Konstantin Schürholt}
\affiliation{%
AIML Lab, University of St. Gallen, Rosenbergstrasse 30, CH-9000 St. Gallen, Switzerland
}

\author{Frank Hellmann}
\affiliation{%
Potsdam Institute for Climate Impact Research, Telegrafenberg A31, 14473 Potsdam, Germany
}%
\email{hellmann@pik-potsdam.de}

\date{\today}

\begin{abstract}
To mitigate climate change, the share of renewable energies in power production needs to be increased. Renewables introduce new challenges to power grids regarding the dynamic stability due to decentralization, reduced inertia, and volatility in production. Since dynamic stability simulations are intractable and exceedingly expensive for large grids, graph neural networks (GNNs) are a promising method to reduce the computational effort of analyzing the dynamic stability of power grids. As a testbed for GNN models, we generate new, large datasets of dynamic stability of synthetic power grids, and provide them as an open-source resource to the research community. We find that GNNs are surprisingly effective at predicting the highly non-linear targets from topological information only. For the first time, performance that is suitable for practical use cases is achieved. Furthermore, we demonstrate the ability of these models to accurately identify particular vulnerable nodes in power grids, so-called \emph{troublemakers}. Last, we find that GNNs trained on small grids generate accurate predictions on a large synthetic model of the Texan power grid, which illustrates the potential for real-world applications. 
\end{abstract}

\maketitle
\begin{quotation}
The future of power grids is shaped by the need to adapt to and mitigate climate change, requiring a transition to carbon-neutral systems with a significant contribution from decentralized solar and wind generators. However, these renewable sources pose challenges to the dynamic stability of power grids due to their lessened ability to respond to power imbalances and frequency deviations.  In this article, we push the limits of a previously introduced approach for assessing the dynamic stability of power grids based on graph neural networks (GNNs). As a prerequisite, we conduct extensive dynamical simulations to generate a considerably larger training dataset. By training larger models on more data we demonstrate that GNNs can accurately predict probabilistic measures of stability. With moderate margins for statistical error, our new GNN models produce stability outcomes more than 1000 times faster than conventional approaches based on dynamical simulations. Furthermore, they generalize from small training grids to a real-world-sized, synthetic model of the Texan power grid. Our study on the prediction of dynamic stability in synthetic power grids takes a crucial step toward deploying machine learning for power grid operation and planning. Future work can build on this and enhance the complexity of power grid models to enable real-world applications. We are convinced that GNNs can have a huge impact for the development of sustainable and resilient power grids in the face of climate change.

\end{quotation}

\section{Introduction}
Adaption to and mitigation of climate change jointly influence the future of power grids: 
1) Mitigation of climate change requires power grids to be carbon-neutral, with the bulk of power supplied by solar and wind generators. These are more decentralized, and as opposed to conventional turbine generators they have less inertia, meaning that there is no intrinsic ability to respond to power imbalances and frequency deviations. Furthermore, the production of renewables is more volatile. Renewable energies will have to contribute to the dynamical stability of the system \citep{milanoFoundationsChallengesLowInertia2018,christensenHighPenetrationPower2020} in the future, requiring a new understanding of the complex synchronization dynamics of power grids.
2) A higher global mean temperature increases the likelihood as well as the intensity of extreme weather events such as hurricanes or heatwaves \citep{fieldManagingRisksExtreme2012,portnerIPCC2022Climate2022}, which result in great challenges to power grids. 
Tackling climate change in the power grid sector calls for building sustainable grids, as well as increasing the resilience of existing power grids toward novel threats at the same time. This requires new methods of understanding and managing dynamic stability. \looseness-1

\paragraph{Introduction of power grids} Power grids are complex networks, consisting of nodes that represent different producers and consumers, as well as edges that represent power lines and power transformers. In contrast to many other networks, the interaction of nodes through the edges is governed by physical equations, the power flow. Their emergent properties can be highly unintuitive; for example, the Braess paradox describes the phenomenon that adding lines to a power grid may reduce its stability \citep{witthautBraessParadoxOscillator2012, schaferUnderstandingBraessParadox2022}. Such effects can be non-local; i.e., the parts of the grid with decreased stability might be far away from the added line. Similarly, failures of a line in one part of the network can lead to overloads far away. Our work deals with the challenge of predicting the ability of the grid to dynamically recover after perturbations.\looseness-1
\paragraph{Dynamics of power grids}
Classically, the dynamical actors are connected to the highest voltage level, the transmission grid. Due to computational bounds, power grid operators are limited to analyze individual faults (called contingencies) at the highest voltage level only without explicitly modeling lower voltage layers of the grid. As distributed renewable generation is typically connected at lower grid levels, this problem will become more acute as renewables start playing a larger role in the grids' dynamics. Conducting high-fidelity simulations of the whole hierarchy of the power grid and exploring all states will not be feasible \citep{liemannProbabilisticStabilityAssessment2021}. For future power grids, knowledge of how to design robust dynamics is required. This has led to a renewed interdisciplinary interest in understanding the collective dynamics of power grids \citep{brummittTransdisciplinaryElectricPower2013}, with a particular focus on the robustness of the self-organized synchronization mechanism underpinning the stable power flow \citep{rohdenSelfOrganizedSynchronizationDecentralized2012,motterSpontaneousSynchronyPowergrid2013,dorflerSynchronizationComplexOscillator2013,witthautCollectiveNonlinearDynamics2022}. Synchronization refers to the fact that a stable power flow requires all generators to establish a joint frequency. It is self-organized in the sense that this is achieved without further communication or an external signal. Synchrony is crucial for power grids, as a stable power flow is only possible if all generators operate at the same frequency. At the same time, the ever present fluctuations in demand have to be compensated by locally varying the frequency. 
\paragraph{Probabilistic modelling of dynamics}
To understand which structural features impact the self-organized synchronization mechanism, it has proven fruitful to take a probabilistic view \citep{menckHowBasinStability2013, menckHowDeadEnds2014, hellmannSurvivabilityDeterministicDynamical2016}. Probabilistic approaches are well established in the context of static power flow analysis \citep{borkowskaProbabilisticLoadFlow1974}. In the dynamic context, considering the probability of systemic failure following a random fault effectively averages over the various contingencies. Such probabilities are, thus, well suited to reveal structural features that enhance the system robustness or identify vulnerable grid regions \citep{menckHowDeadEnds2014, schultzDetoursBasinStability2014, nitzbonDecipheringImprintTopology2017,hellmannNetworkinducedMultistabilityLossy2020}. Recently, probabilistic stability assessments gained more attention in the engineering community as well \citep{liuQuantifyingTransientStability2017,liuBasinStabilityBased2019,liemannProbabilisticStabilityAssessment2021}.\looseness-1

\paragraph{Application of Machine Learning}
Given the need for probabilistic analysis and the computational cost of explicit simulations, we apply graph neural networks (GNNs) to directly predict probabilistic measures from the system structure. Such GNNs could be used to screen many potential configurations and select critical ones for which a more detailed assessment should be carried out. Moreover, the analysis of the decision process of ML models might lead to new unknown relations between dynamical properties and the topology of grids. Such insights may ultimately inform the design and development of power grids.

Since datasets of probabilistic stability in power grids of sufficient size do not exist yet, we introduce new datasets, which consist of synthetic models of power grids and statistical results of dynamical simulations that required roughly 700,000 CPU hours. We simulated datasets of increasing complexity to get closer to reality step by step. There are 10 000 small grids, 10 000 medium-sized grids, and, for evaluation purposes, one large grid based on a synthetic Texan power grid model.\looseness-1

\paragraph{Related work on power grid property prediction}
Since power grids have an underlying graph structure, the recent development of graph representation learning \citep{bronsteinGeometricDeepLearning2021,hamiltonGraphRepresentationLearning2020} introduces promising methods to use machine learning in this domain. There are a number of applications dealing with GNNs and different power flow-related tasks \citep{dononGraphNeuralSolver2019,kimGraphConvolutionalNeural2019,bolzPowerFlowApproximation2019,retiereSpectralGraphAnalysis2020,wangProbabilisticPowerFlow2020,owerkoOptimalPowerFlow2020,gamaGraphNeuralNetworks2020,misyrisPhysicsInformedNeuralNetworks2020,liuSearchingCriticalPower2021,bushTopologicalMachineLearning2021,liuGuidingCascadingFailure2020,jhunPredictionMitigationNonlocal2022,chenEvaluatingDistributionSystem2022,stoverJustInTimeLearningOperational2022,hansenPowerFlowBalancing2022,yanivAdoptionGNNsPower2023} and to predict transient dynamics in microgrids \citep{yuPIDGeuNGraphNeural2022}. There is also literature using conventional ML methods dealing with the basin stability \citep{cheActiveLearningRelevance2021,yangPowergridStabilityPredictions2021} in the context of power grids. \citet{nauckPredictingBasinStability2022} use small GNNs to predict the dynamic stability on small datasets. They demonstrate the general feasibility of the approach, but do not compare to conventional baselines. We add such baselines and introduce larger datasets to train GNNs with much higher capacity to achieve better predictive power. 

\begin{figure}
    \centering
    \includegraphics[width=\linewidth]{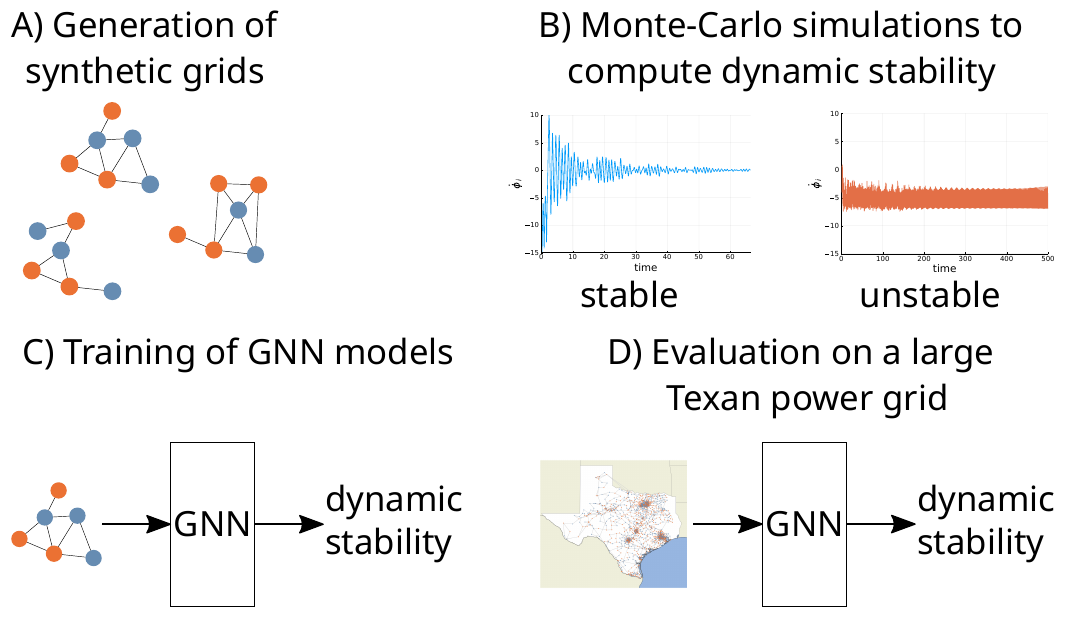}
    \vspace{-20pt}
    \caption{We generate new datasets of the dynamic stability of power grids, based on synthetic power grids (A) and statistics of dynamical simulations (B). Then, GNN models are trained to predict the dynamic stability of the synthetic grids (C) and the models are evaluated on a Texan power grid model (D).\looseness-1}
    \vspace{-11pt}
	\label{fig_scheme}
\end{figure}

\paragraph{Our main contributions are:}
We introduce new datasets of probabilistic dynamic stability of synthetic power grids. The new datasets have ten times the size of previously published ones and include a Texan power grid model to map the path toward real-world applications. We also observe a relevant new class of nodes:  so-called \emph{troublemakers}, at which perturbations are strongly amplified. Such nodes may be dangerous to hardware and the overall grid stability. Their identification constitutes an additional task. We train strong baselines and benchmark models to evaluate the difficulty of all tasks. Our results demonstrate i) that the larger dataset allows training more powerful GNNs, (ii) which outperform the baselines, and (iii) transfer from the new datasets to a real-sized power grid. Using larger datasets and better models, the performance reaches levels that may become relevant for real-world applications for the first time. The general approach is visualized in \Cref{fig_scheme}.\looseness-1

\section{Generation of the datasets}
\label{sec_generation_dataset}

\subsection{Modeling power grids as dynamical systems}
Full scale analysis that can treat high-fidelity models of real systems is currently out of reach for several reasons. These include that real-world data do not exist or are not accessible, synthetically generating large numbers of realistic grids is challenging, and that large dynamical models cannot be simulated fast enough with current software \citep{liemannProbabilisticStabilityAssessment2021}. These problems force trade-offs on us, most notably reducing the details of the intrinsic behaviors of dynamical actors to that of inertial oscillators. This leads to the Kuramoto model~\cite{kuramotoSelfentrainmentPopulationCoupled1975}, a paradigmatic model for synchronization studies \citep{kuramotoSelfentrainmentPopulationCoupled2005,rodriguesKuramotoModelComplex2016}, which was introduced for power grids in \citep{bergenStructurePreservingModel1981}. This model strikes a careful balance of capturing the key dynamics that govern synchronization in real power grids while remaining computationally--and to some degree analytically--tractable. It is thus highly useful for understanding the relationship between the grid structure and synchronization, but it should not be taken to be a fully adequate model of the grid by itself. Any future treatment of dynamic stability based on more accurate models will also have to solve the challenging sub-problem of the impact of topology on synchrony that we consider here; thus, we see this work as an important first step.\looseness-1

In an ideal AC grid, the nodes' voltage is given by a 50 or 60~Hz sine curve. Writing $\phi_i$ for the time-dependent deviation of the phase of the local AC voltage from an arbitrary reference signal at node $i$, the instantaneous power flow on the line from node $i$ to node $j$ at time $t$ is given in terms of the line parameter $K$ by $P_{ij}(t) = K \sin(\phi_i(t) - \phi_j(t))$. The normalized time derivative $\frac{1}{2\pi}\dot \phi_i$ provides the local frequency deviation from the reference frequency. The two most important dynamical processes that establish synchrony and a stable power flow are I) inertia, i.e., the change of local frequency as a result of absorbing local power imbalance $M \ddot \phi_i = \Delta P_i$, where $M$ is the inertia and II) droop control, the local change of injected power due to frequency deviation $P_i^\mathrm{droop} = - \alpha \dot \phi_i (t)$, using the droop parameter $\alpha$.

In order to extract the impact of topology as cleanly as possible, we assume homogeneous inertia constant $M$, droop parameter $\alpha$, and line parameters $K$. The dynamical equations for the self-organized synchronization and stabilization of the active power flow are then given by conservation of energy. Let $P^d_i$ denote the power injected/consumed at node $i$ and $A_{ij}$ the adjacency matrix:
\begin{align}
   \Delta P_i &= P^d_i + P_i^\mathrm{droop} - \sum_j A_{ij} P_{ij}(t),\\
   M \ddot \phi_i &= P^d_i - \alpha \dot \phi_i - K \sum_j A_{ij}  \sin(\phi_i(t) - \phi_j(t)). \label{eqKuramoto}
\end{align}

Synchronous operation requires $0 = \ddot \phi_i(t) = \dot \phi_i(t)$. The fixed point equation $P^d_i = K \sum_j A_{ij} \sin(\phi^*_i(t) - \phi^*_j(t))$ is the power flow equation on a grid with topology given by the adjacency matrix $A$.

To generate synthetic power grids, distinctive topological properties have to be considered. Power grids are sparsely connected. The degree distribution has a local maximum at very small degrees (e.g., $\approx2.3$), an exponentially decaying tail, and a mean of $\approx2.8$ \citep{schultzRandomGrowthModel2014}. Hence, most nodes are only connected to a few neighbors, though some highly connected nodes typically do exist. 

\subsection{Quantifying dynamic stability of power grids \label{sec_snbs}}
We quantify dynamic stability with the single-node basin stability (SNBS) \citep{menckHowBasinStability2013}. This measure is widely used in the study of synchronization phenomena. As mentioned in the introduction, it is probabilistic and defined as the probability that the system recovers following a perturbation by a random fault. 

In the dynamical systems' community, the  expected perturbations are typically modeled by a distribution of initial conditions of the post fault system. For every node $i$, let $\rho_i$ be a distribution of initial conditions corresponding to contingencies localized at that node. For a power grid with $N$ nodes modeled by \Cref{eqKuramoto}, denote the state space trajectory as $(\phi(t), \dot \phi(t)) = (\phi_j(t), \dot \phi_j(t))_{j=1,\dots,N}$, and the fixed point as $ (\phi^*, 0 )$. Then the SNBS at node $i$ is the probability that the system's trajectory returns to the fixed point starting from an initial condition $(\phi(0), \dot \phi(0)) = (\phi_0, \dot \phi_0)$ drawn from $\rho_i$.

This probability can be estimated as the outcome of a Bernoulli experiment. Here, we simulate 10,000 trajectories per node to minimize statistical errors. The underlying simulations to generate the datasets are feasible for anyone with some domain knowledge, but the composition of entire datasets requires significant amounts of computational resources. For the datasets with 20,000 grids and the Texan power grid combined, the simulations take roughly 700,000 CPU hours. Thus, an important contribution is to publish a full dataset to enable groups that have less computational resources to work on this important problem. 
The goal of our work is to replace these expensive simulations by GNNs. To that end, we train GNNs to learn the graph function $ (P, A) \rightarrow \mathrm{SNBS}$ where $P$ is a featurized version of $P^d$.

\subsection{Modeling of the Texan power grid \label{sec_modeling_texas}}
To take a further step toward real-world applications, we evaluate the performance of our GNN models by analyzing the dynamic stability of a real-sized synthetic power grid. Real power grid data are not available due to security reasons. We chose a synthetic model derived from the Texan power grid topology, introduced by \cite{birchfieldGridStructuralCharacteristics2017,birchfieldACTIVSg20002000busSynthetic2021}. The synthetic Texan power grid model consists of 1 910 nodes after removing 90 nodes that are not relevant for the dispatching. We use the same modeling approach as for the other grids, i.e, we use only the topological properties. As a consequence, we only investigate the potential applicability of GNNs to real-sized grids and cannot make any statements about the real-world Texan power grid. Even after applying the simplifications, the simulations are already very expensive due to a large number of nodes. To manage the computational cost of simulating dynamic stability of such a large grid, we reduce the number of simulated perturbations from 10 000 to 1 000. Nonetheless, the simulation of that grid takes 127 000 CPU hours. Computing less perturbations results in an increased standard error  of approximately $\pm 0.031$ for the SNBS estimates.  \looseness-1

\subsection{New troublemaker definition}
\label{sec_method_trouble_makers}
\begin{figure}
    \centering        \includegraphics[width=.99\linewidth]{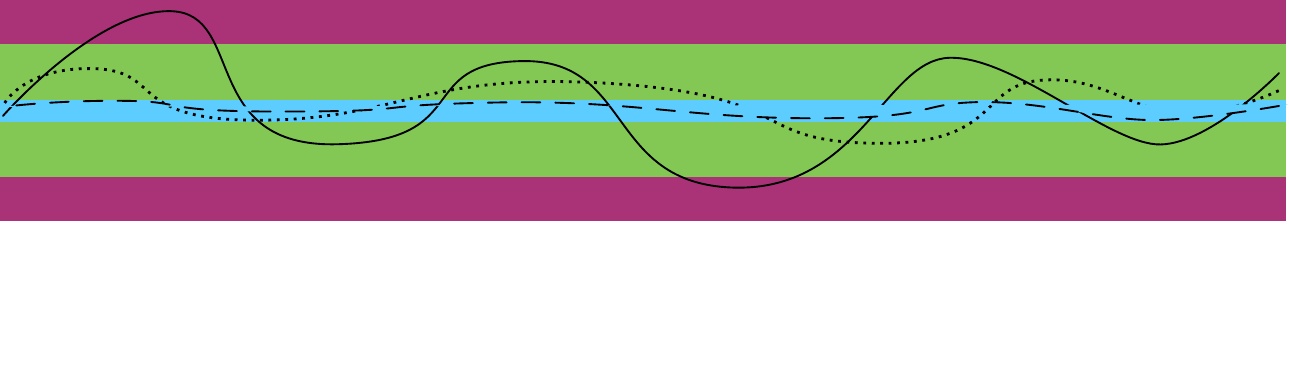}
    \vspace{-30pt}
    \caption{Identification of \emph{troublemakers} based on trajectories. There are three spaces: \textcolor{blue_perturbation_space}{blue} for the space of perturbations, \textcolor{green_stable}{green} for the \emph{safe}-space and \textcolor{purple_trouble}{purple} for the \emph{trouble}-space. The dashed trajectory stays within the initial space of the perturbations, the dotted line leaves that space, but stays below the \emph{trouble}-threshold and the solid line represents a trouble-maker-trajectory.}
 	\vspace{-11pt}
	\label{fig_scheme_trouble_makers}
\end{figure}

We introduce a new category for nodes called \emph{troublemakers} that amplify  perturbations by a large factor. Previously, we looked at SNBS that considers the asymptotic stability, which is not sufficient to ensure stable grid operation at all times. If the transient trajectories after the perturbation violate operational bounds, machines in the system switch off to protect themselves, potentially triggering failure cascades and large blackouts. This motivates the definition of survivability \citep{hellmannSurvivabilityDeterministicDynamical2016}, the probability that the system stays within these bounds after a fault. In the generated datasets, we observe that there are some nodes for which the transient reaction is vastly larger than the initial perturbation and call such nodes \emph{troublemakers}, because of the danger they pose to stable grid operation. The idea is illustrated in \Cref{fig_scheme_trouble_makers}. The word troublemakers was previously used by \citet{auerStabilitySynchronyLocal2017}  for a related, but different concept in power grids with persistent fluctuations due to renewable energy sources.

We define the maximum frequency deviation (mfd) of the whole system from an initial condition $(\phi_0, \dot \phi_0)$ as the maximum over all times $t$ and all nodes $j$,

\begin{equation}
    \mathrm{mfd}(\phi_0, \dot \phi_0) := \max_{t, j} \left| \left\{  \dot \phi_j(t)  \mid  \phi(0) = \phi_0,  \dot \phi(0) = \dot \phi_0 \right\} \right|.
\end{equation}

For a given distribution of initial conditions $\rho$, we define

\begin{equation}
    \mathrm{MFD}(\rho) := \max \left\{ \mathrm{mfd}(\phi_0, \dot \phi_0) \mid (\phi_0, \dot \phi_0) \in \mathrm{supp}\,\rho \right\}
\end{equation}
as the maximum frequency deviation caused by any possible perturbation according to $\rho$. When dealing with a fixed
 distribution of localized contingencies at a node~$i$, denoted $\rho_i$, we use the simplified notation $\mathrm{MFD}_i := \mathrm{MFD}(\rho_i)$. If we knew $\mathrm{MFD}_i$ exactly for every node, it would be very straightforward to classify nodes as safe or unsafe: Just check that $\mathrm{MFD}_i$ is below the critical frequency threshold considered secure. Unfortunately, there are no algorithms for computing $\mathrm{MFD}_i$ exactly, and even estimating it statistically is unreliable due to the maximum involved in its definition. To make the important concept of maximal frequency deviations accessible with statistical estimation techniques, its definition may be slightly relaxed.
We may consider nodes as safe if they have a very small probability $\gamma$ that the $\mathrm{mfd}$ of a fault drawn from $\rho_i$ is larger than a critical threshold $\beta$, and other nodes are troublemakers (TMs). This is an expectation value that can be estimated. Writing $\left[\mathrm{mfd}(\cdot) < \beta \right]$ for the function that is~$1$ if $\mathrm{mfd}(\cdot) < \beta $ and $0$ otherwise, we have
\begin{align}
\label{eq:TM max freq dev}
\begin{split}
\mathrm{TM}_i = \left\{
    \begin{array}{lr}
        0 & \mathrm{if } \ \mathbb{E}_{\rho_i}\left[\mathrm{mfd} < \beta \right] >  1 - \gamma, \\
        1 & \mathrm{otherwise}.
    \end{array}
\right.
\end{split}
\end{align}

Importantly, the expected value $\mathbb{E}_{\rho_i}\left[\mathrm{mfd} < \beta \right]$ is the success probability of a Bernoulli experiment and can be estimated from simulations. Given $n$  trials of which $s$ are successes, the success probability $p$ of a Bernoulli random variable may be estimated as $\widehat p = \frac{s}{n}$. Since underestimating the probability of failures  might have grave consequences for power grids, our definition is not based on the point estimate $\widehat p$ , but on the lower bound $p^-_{1-\alpha}(n,s)$ of the confidence interval for the probability of seeing $s$ successes in $n$ trials  at confidence level $1-\alpha$. In this work, we choose a relatively strict confidence level of $0.999$ (that is $\alpha = 0.001$). This means, that if the true success probability would be less than $p^-_{0.999}(n,s)$, the chance of seeing at least $s$ successes in $n$ trials is less than $0.1 \%$. 

Then, given $n_i$ simulations at node $i$, out of which $s_i$ stay within the operational bounds, the empirical estimator of $TM_i$ is 
\begin{align}
\label{eq:TMhat}
\begin{split}
\widehat{\mathrm{TM}_i} = \left\{
    \begin{array}{lr}
        0 & \mathrm{if } \ p_{0.999}(n_i,s_i) \geq 1- \gamma  \\
        1 & \mathrm{else}.
    \end{array}
\right.
\end{split}
\end{align}
In general, we would like to choose $\gamma$ as small as possible to minimize the failure probability. In practice, we choose it in relation to the number of available samples. The number of samples determines the maximum possible value of $p^-_{1-\alpha}$. Note that, the probability of seeing $n$ successes in $n$ trials is $p^n$. Requiring $p^n \geq \alpha$, we get $p^-_{1-\alpha}(n,n) = \sqrt[n]{\alpha}$;
for example, if we have 1000 samples, then $p_{0.999}(n,s) \leq p_{0.999}(n,n) = \sqrt[n]{0.001} \approx 0.9931$.
 In this case, choosing $\gamma = 0.001$ would lead to all nodes being classified as troublemakers--even if in the 1000 trials no failures had been observed. A more reasonable value would be $\gamma = 0.01$.

There are many methods of computing confidence intervals for binomial proportions like $\mathbb{E}_{\rho_i}\left[\mathrm{mfd} < \beta \right]$. In this work, we use a one-sided Clopper-Pearson interval because its empirical coverage is guaranteed to be more than $1-\alpha$~\cite{brownIntervalEstimationBinomial2001}. It is defined as
\begin{equation}
    p^-_{0.999}(n,s) :=  \inf \left\{p \mid \mathbb{P}\left[Bin\left(n,p\right) \geq s \right] > 0.001 \right\}.
\end{equation}

For choosing the critical threshold $\beta$, consider that the outer limits for frequency deviations for the European grid are $+2$Hz or $-3$Hz. For our model, we choose the symmetric limit of $\approx2.4$Hz, both for simplicity and comparability to previous work. In the Kuramoto model, this corresponds to a maximum deviation of the angular velocity of $|\dot \theta| < 2\pi \cdot 2.4 \approx 15 =: \beta$.

To focus on the worst offenders, in \Cref{sec:id_tm} we predict nodes that amplify perturbations by at least a factor of $6$. In our power grid models, this corresponds to initial perturbations confined to $[-2.5, 2.5]$ of the angular velocity. In the terms introduced above, the localized measure $\rho_i$ of perturbations at a node $i$ is chosen to be the uniform distribution on $[-2.5, 2.5] \times [-\pi,\pi)$, while the desirable region is $[-15, 15] \times [-\pi,\pi)$ at all nodes. The observation of large amplifications and the target function are both of practical importance and novel as far as we are aware. Details regarding the error bounds are given in \Cref{app_TM_error_bounds}. \looseness-1

\subsection{Structure of the datasets \label{sec_structure}}
To generate the datasets, we closely follow the methods in \cite{nauckPredictingBasinStability2022} and extend their work by computing ten times as many grids. To investigate different topological properties of differently sized grids, we generate two datasets with either 20 or 100 nodes per grid, referred to as dataset20 and dataset100. To enable the training of complex models, both datasets consist of 10,000 graphs. Additionally, one large synthetic Texan grid is provided for testing out-of-sample performance; see \Cref{sec_modeling_texas}.

For every grid, two input features are given, namely the adjacency matrix $A \in \{0,1\}^{N \times N}$ representing the topology and a binary feature vector $P^d_i \in \{-1,1\}^{N}$, specifying the power injection/demand at the nodes in  \Cref{eqKuramoto}. Here, $N$ is the number of nodes. Likewise, for every grid, the SNBS target vectors are given: $\mathrm{SNBS} \in \left[0,1\right]^N$. Furthermore, we provide the target vector $\mathrm{TM} \in \{ 0,1 \}^N$ (classification) for each grid. The TM class is derived from the maximum frequency deviation $\mathrm{MFD} \in \left[0,\infty \right)^N$, which is also provided in the dataset. Since the $\mathrm{MFD}$ values are derived from the maximum frequency derivation of the sampled trajectories, no error bounds can be given. However, MFD can still be used as targets, when applying thresholding after training to predict troublemakers; see \Cref{sec_method_trouble_makers}. 

\begin{figure*}[t]
    \centering
    \vspace{-20pt}
        \includegraphics[width=\textwidth]{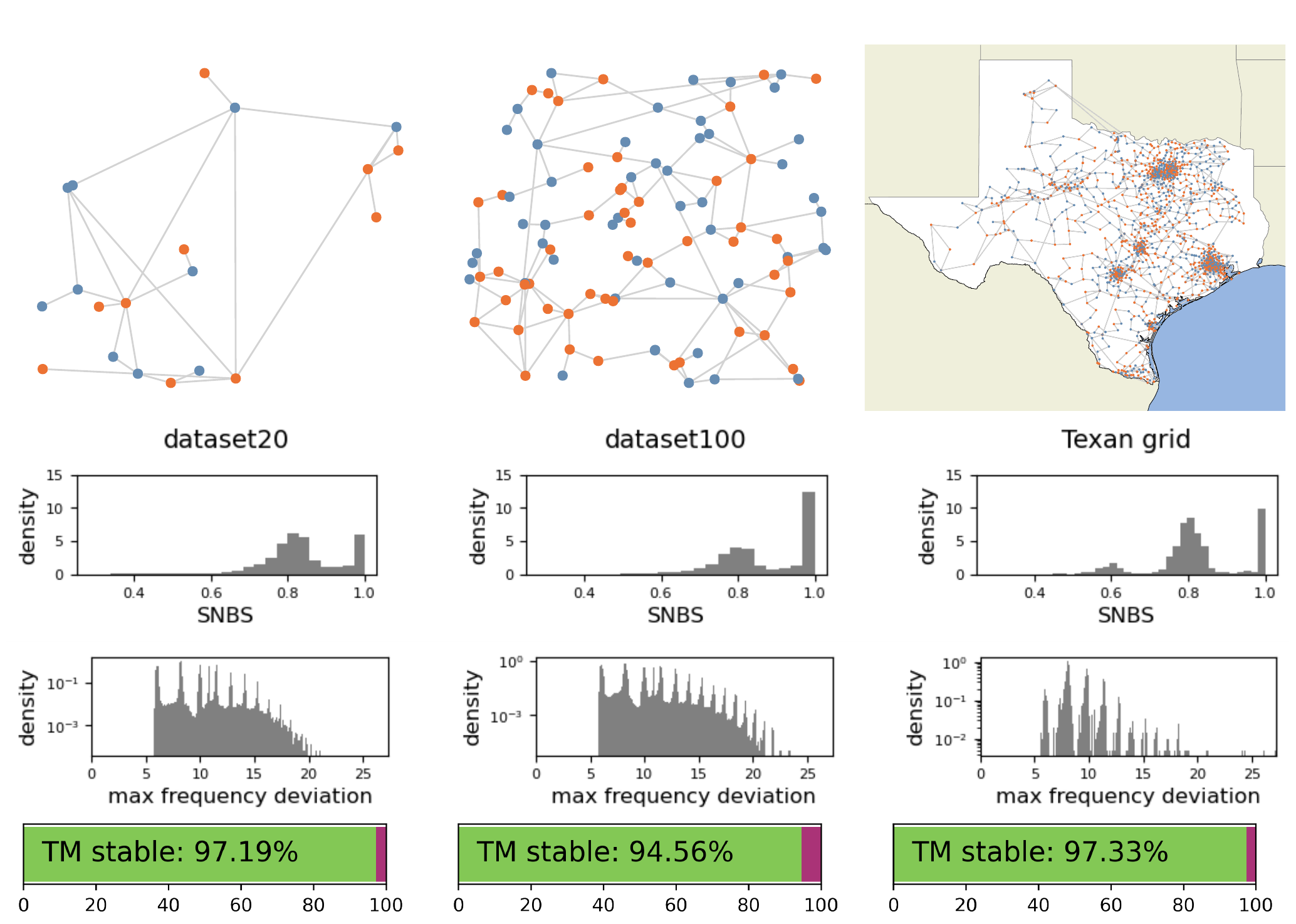}
    \vspace{-11pt}
    \caption{Examples of power grids in the datasets with 20 nodes (top left) and 100 nodes (top center) and the Texan power grid model (top right). \textcolor{blue_source}{Blue} color denotes sources and the \textcolor{orange_sink}{orange} sinks. Below, the histograms of SNBS and the maximum frequency deviations (logarithmic scale) are shown. At the bottom, the share of troublemakers (TM) is shown, where \textcolor{green_stable}{green} represents stable nodes and \textcolor{purple_trouble}{purple} troublemakers (TM).}
	\label{fig_gridExamples_distributions}
\end{figure*}

Examples of the grids of dataset20, dataset100 and the Texan power grid as well as the distributions of SNBS (characterized by multiple modes) and the troublemaker nodes TM (imbalanced binary classification task derived from the MFD) are given in \Cref{fig_gridExamples_distributions}. Even though the same modeling approach is used, there are significant variations that are entirely caused by different topologies and grid sizes. Interestingly, the SNBS distribution of the Texan power grid has a third mode, which is challenging for prediction tasks. Overall, the power grid datasets consist of the adjacency matrix encoding the topology, the binary injected power $P$ per node as input features, and nodal SNBS, MFD, and TM. \looseness-1

\section{Predicting dynamic stability of power grids using GNNs}
In this section, we predict the dynamic stability of the new datasets using GNNs. We start by introducing GNNs, followed by the experimental setup, results of predicting SNBS on the two grid sizes, as well as analyzing out-of-distribution capabilities from small to large grids. Subsequently, we establish the advantages of our large dataset in comparison with the previous work. Afterward, we evaluate the GNNs trained on our datasets on the Texan power grid as a larger, more realistic test-case. Last, we identify troublemakers. \looseness-1

\subsection{Theoretical background on graph neural networks}
Graph neural networks are a class of artificial neural networks designed to learn relationships of graph-structured data. They have internal weights, which can be fitted in order to adapt their behavior to the given task. GNNs use the graph structure as inputs and potentially node and edge features, too. Their output can either be global graph attributes, attributes of sub-graphs, or local node properties. Different types of GNN have been introduced and many of the layers build on the graph convolution network (GCN) introduced by \citet{kipfSemiSupervisedClassificationGraph2017}:
\begin{equation}
	H = \sigma(\overline{A} X \Theta),
\end{equation}
where $H$ is the output of a layer, $\sigma$ denotes the activation function, using the input features $X$, the matrix $\Theta$ containing the learnable weights, and a slightly modified and re-normalized adjacency matrix $\overline{A}$. To increase the considered region and to consider neighbors at further distance, multiple GCN layers can be applied consecutively. Recently, more complex GNN layers have been developed to allow more complex graph convolutions, but the basic idea of aggregating information on a graph structure has not changed. 

\subsection{Experimental setup}
\label{sec_Experimental_setup}
\begin{figure}
    \centering
    \includegraphics[width=\linewidth]{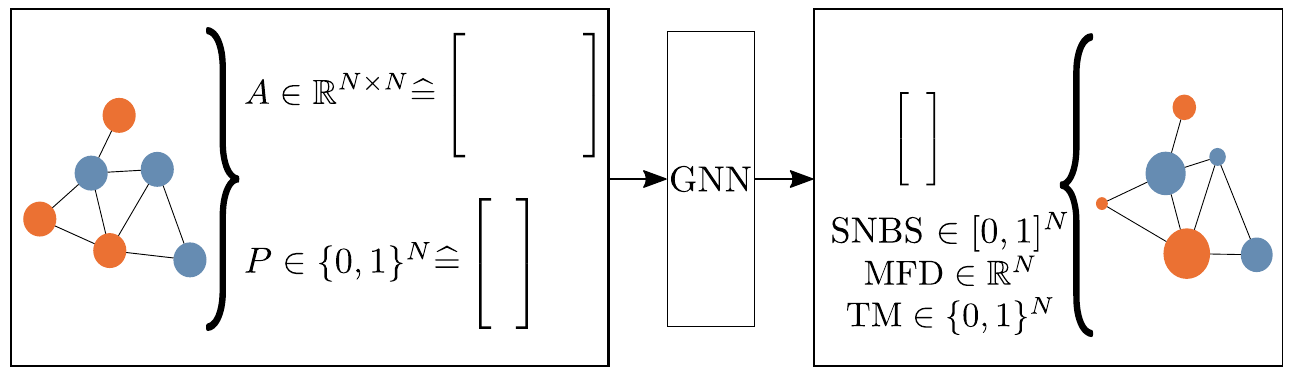}
    \vspace{-11pt}
    \caption{Prediction of nodal outputs SNBS and TM using GNNs. The inputs are the adjacency matrix $A$ and the injected power $P^d$. The prediction is purely based on the structure and topology of the grid and does not consider individual faults.}
    \vspace{-12pt}
    \label{ProcessPrediction}
\end{figure}

We train GNNs on nodal prediction tasks, using regression for SNBS, and regression with thresholding in the case of MFD as well as classification for TM (visualized in \Cref{ProcessPrediction}). As input, the GNNs are given an adjacency matrix and the power injection/demand at the nodes; cf. \Cref{sec_structure}. We split the datasets in training, validation, and testing sets (70:15:15). To minimize the effect of initialization, we use five different initializations per model and compute average performances using the three best ones.

To evaluate the robustness of the GNNs, we analyze the performance of different models based on several GNN architectures: GNNs with ARMA filters by \citet{bianchiGraphNeuralNetworks2021}, graph convolutional networks (GCN) by \citet{kipfSemiSupervisedClassificationGraph2017}, SAmple and aggreGatE (SAGE) by \citet{hamiltonInductiveRepresentationLearning2018} and topology adaptive graph convolution (TAG) by \citet{duTopologyAdaptiveGraph2017}. We refer to the models by ArmaNet, GCNNet, SAGENet and TAGNet. We conduct hyperparameter studies to optimize the model structure regarding number of layers, number of channels and layer-specific parameters using dataset20. Afterward, we optimize learning rate, batch size, and scheduler for dataset20 and dataset100 separately. Details on the hyperparameter study and the models are given in \Cref{sec_hyperparameter}.\looseness-1

\paragraph{Baseline models}
To better assess the GNN performance, we set up several baseline models. \citet{schultzDetoursBasinStability2014} were the first to attempt predicting nodal dynamic stability of power grids using a logistic regression of common network measures and hand-crafted features. For the first baselines, we set up similar linear regression models with the following input features: degree, average-neigbor-degree, clustering-coefficient, current-flow-betweenness-centrality, closeness-centrality and the injected power $P^d$. Additionally, we use more complex Multilayer perceptrons (MLPs) trained on the same features as baselines. We investigate the performance of two differently sized MLPs, where MLP1 has 1 541 parameters and MLP2 1 507 001 parameters; see \Cref{appendix_MLP} for details.
\tikzstyle{output}=[very thick, minimum size=1.5em, draw=white!100, fill=white!100, minimum width=2.5em, minimum height = 5em]
\tikzstyle{layer}=[very thick, minimum size=1.5em, draw=black!100, fill=white!100, minimum width=2.5em, minimum height = 6em]

\paragraph{Metrics for evaluation}
To analyze the performance, we use the coefficient of determination ($R^{2}$-score) for regression and F2-score for classification. F2-score is a modified F-score giving more weight to recall and less to the precision in the calculation of the score. In the case of identifying vulnerabilities of power grids, it is more important to identify all critical states, even if this increases the number of false positives. The details are provided in \Cref{sec_evaluation_metrics}.
\subsection{Prediction of SNBS}
\begin{figure*}
    \centering
        \includegraphics[width=\linewidth]{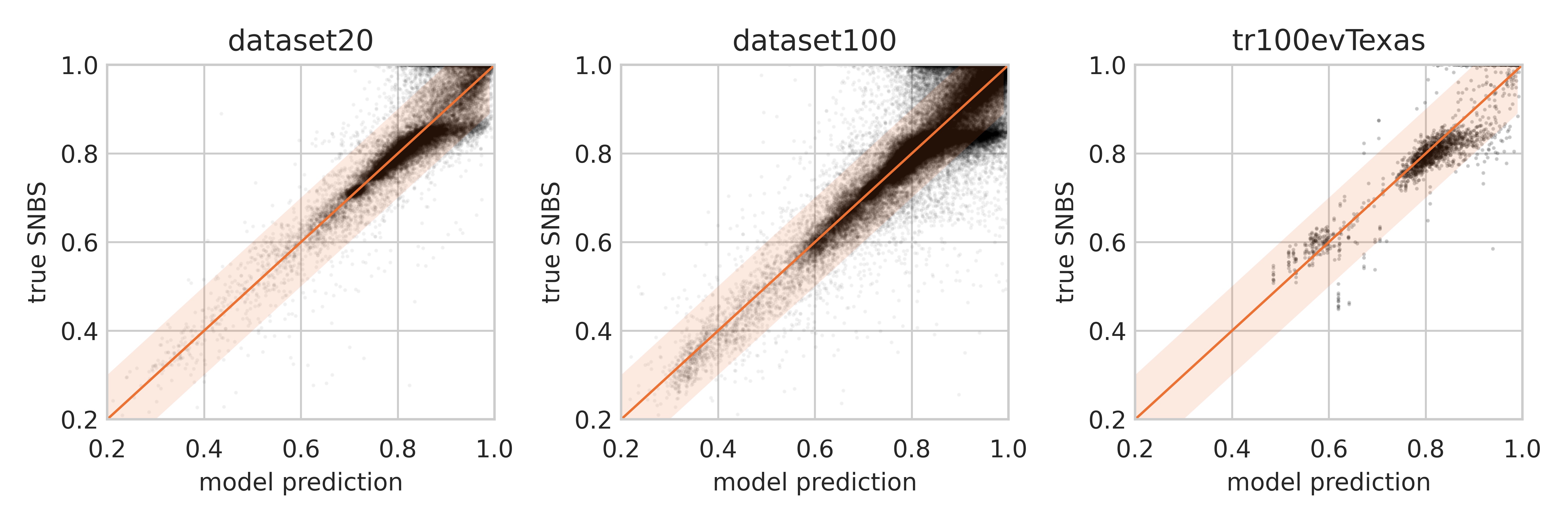}
    \vspace{-10pt}
    \caption{SNBS over predicted output of the TAGNet model for the in-distribution tasks on dataset20 and dataset100, and the out-of-distribution generalization from dataset100 to the Texan power grid (tr100evTexas). The diagonal represents a perfect model ($R^2 = 1$), the banded region indicates predictions which are accurate up to an error of $\pm0.1$.}
    \vspace{-11pt}
	\label{fig_result_scatter_dataset_comparison}
\end{figure*}

\paragraph{GNNs can accurately predict SNBS}
In our first set of benchmark experiments, the goal is to predict SNBS with high accuracy, as measured by the coefficient of determination $R^2$. The key result is the surprisingly high performance of GNNs across all datasets, see the first two columns in \Cref{tb_results_snbs}. $R^2$ reaches values above 82 \% for dataset20 and above 88 \% for dataset100. SNBS is a highly nonlinear property, and the obtained performance exceeds expectations. The predictive performance captures not only the general trend, but the modalities in the data as well (cf. \Cref{fig_result_scatter_dataset_comparison}). Interestingly, the previously published Ar-bench \citep{nauckPredictingBasinStability2022} performs worse than the MLPs, but the larger GNNs outperform all baselines.\looseness-1

\begin{table*}[htb]
    \small
	\centering
	\caption{Results of predicting SNBS represented by $R^2$ score in \%. Each column represents the evaluation on a different test set, e.g. \textit{tr20ev20} denotes that the models are trained and evaluated using dataset20. Additionally, we analyze the out-of-distribution capabilities by evaluating the models on different datasets without retraining, e.g., we train a model on dataset20 and evaluate it on dataset100 and refer to this by \textit{tr20ev100}. Besides the performance of the GNNs, we show the performance of a linear regression and multilayer perceptrons using hand-crafted features as baselines.}
	\begin{tabularx}{\linewidth}{XXXXXX}
		\toprule
	   model & tr20ev20 & tr100ev100 & tr20ev100 & tr20evTexas & tr100evTexas \\
        Ar-bench & 52.16 \tiny{$\pm$ 3.36} & 60.67 \tiny{$\pm$ 0.28} & 36.75  \tiny{$\pm$ 1.49} & 46.74 \tiny{$\pm$ 2.31} &  58.36 \tiny{$\pm$ 3.62}\\
        ArmaNet &  82.22 \tiny{$\pm$ 0.12}&  \textbf{88.35} \tiny{$\pm$ 0.12} & \textbf{67.12} \tiny{$\pm$ 0.80} & 52.50 \tiny{$\pm$ 2.68} & 63.95 \tiny{$\pm$ 2.27}\\
        GCNNet  & 70.74 \tiny{$\pm$ 0.15} & 75.19 \tiny{$\pm$ 0.14} & 58.24  \tiny{$\pm$ 0.47} & 50.17 \tiny{$\pm$ 3.60} & 48.56 \tiny{$\pm$ 1.02} \\
        SAGENet & 65.65 \tiny{$\pm$ 0.13} &  75.44 \tiny{$\pm$ 0.33} &  52.30  \tiny{$\pm$ 0.35} & 33.45 \tiny{$\pm$ 1.00}  & 54.38 \tiny{$\pm$ 0.41}\\
        TAGNet & \textbf{82.50} \tiny{$\pm$ 0.36} & 88.32 \tiny{$\pm$ 0.10} & 66.32  \tiny{$\pm$ 0.74}&  \textbf{58.43} \tiny{$\pm$ 1.25} & \textbf{83.31} \tiny{$\pm$ 1.46} \\
        \midrule
        linreg & 41.75 & 36.29 & 5.98 & -11.44 & -22.63 \\
        MLP1 & 59.07  \tiny{$\pm$ 0.04 } & 66.30  \tiny{$\pm$ 0.05}  & 29.98 \tiny{$\pm$ 3.72} & -13.22 \tiny{$\pm$ 22.07} & -5.13  \tiny{$\pm$ 18.00}\\
        MLP2 &  56.59 \tiny{$\pm$ 0.07}&  63.66 \tiny{$\pm$ 0.08 } & 33.09 \tiny{$\pm$ 0.55} &  -4.62\tiny{$\pm$ 5.01} &  19.58 \tiny{$\pm$ 1.59}\\
	 \bottomrule
 	\end{tabularx}
 	\label{tb_results_snbs}
\end{table*}

\paragraph{GNNs generalize from dataset20 to dataset100}
Note that, as reviewed in the introduction, the dynamical properties of the power grid are non-linear and non-local. Our perturbations are localized, but the probabilistic measures look at the whole system response. Thus, it is \textit{a priori} unclear how well we can expect the models to generalize from smaller grids to larger ones. Training on small grids without loss of generalization and predictive power would be a huge advantage to scale to real power grids. To evaluate the potential of our datasets and GNN models to that end, we apply an out-of distribution task by training the models on dataset20 and evaluating the performance without any further training on dataset100. The third column in \Cref{tb_results_snbs} (tr20ev100) shows that most GNNs generalize well, and the best are able to predict SNBS with $R^2$ exceeding 66 \%. We would like to emphasize the significance of that finding. Given sufficient size and complexity in the source dataset, GNNs can robustly predict highly nonlinear stability metrics for grids several times larger than the source. We did not expect grids of size 20 to be large enough to contain enough relevant structures to generalize to larger grids. Generalizing from small, numerically solvable grids to large grids is key for real-world application. The computational cost of the dynamic simulations scales at least quadratic with the size of the grids; therefore computational time can be saved when training models on smaller networks or sections of real-sized grids. In comparison with the baselines, the generalization capabilities of the new GNN models are much better.\looseness-1

\begin{figure}[h]
    \centering
    \vspace{-10pt}
     \includegraphics[width=.95\linewidth]{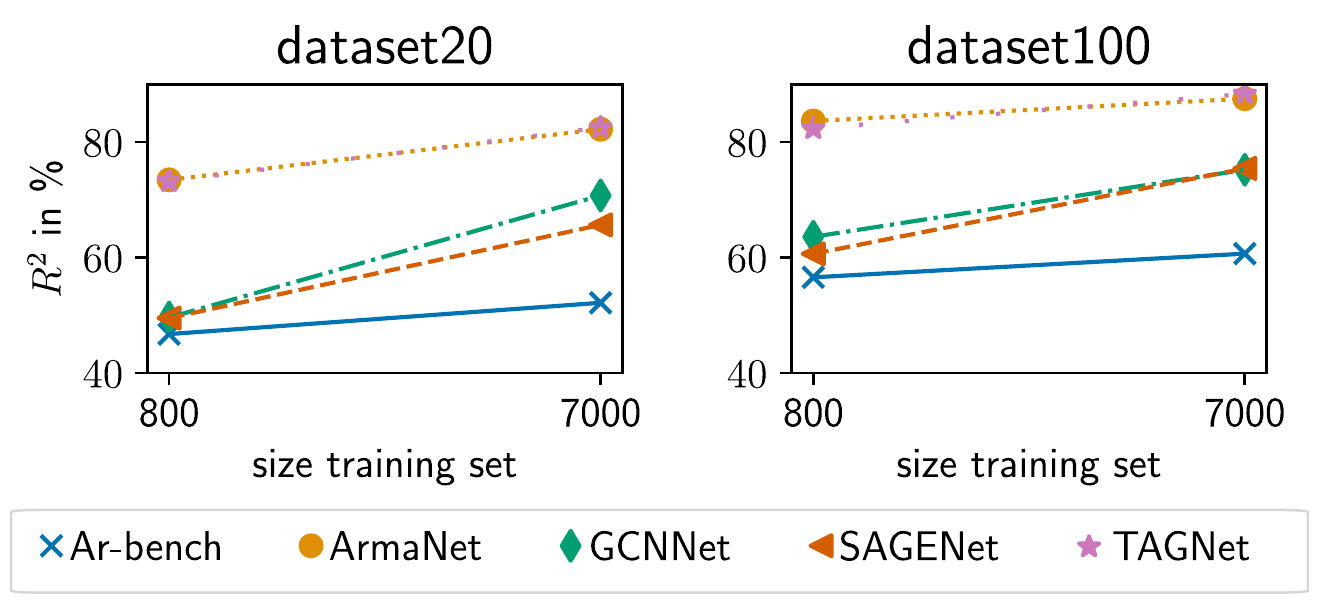}
    \vspace{-10pt}
    \caption{Comparison of the performance based on the size of the training set using 800 or 7 000 grids. Training on our larger dataset improves performance on all models. The 800 grids used for the training follow \cite{nauckPredictingBasinStability2022} and all models are evaluated on a newly introduced test set.
    }
    \vspace{-11pt}
	\label{fig_more_data_better_performance}
\end{figure}

\paragraph{Training on more data increases the performance of all models} 
General machine-learning convention assumes that larger dataset size allows training larger models to higher performance. In this section, we investigate the influence of the size of the training set to show the relevance of the larger datasets. We train the models on the smaller dataset introduced in \cite{nauckPredictingBasinStability2022} after specifically optimizing the learning rate for the analysis of a smaller training set. Our experiments show that training on fewer data results in lower performance; see \Cref{fig_more_data_better_performance}. Instead of peak values of $R^2$ of 82.49 \%, we only obtain 74.77 \% for dataset20 and only 83.92 \% instead of 88.22 \% for dataset100. The results of all models are given in \Cref{sec_appendix_smaller_dataset}. Comparing the performance differences on dataset20 and dataset100, the improvements are larger for dataset20. A reasonable explanation is the total number of nodes used for the training.

\paragraph{Predicting SNBS on a Texan power grid model using the previously trained models \label{sec:Texas}}
Using GNNs for SNBS prediction becomes feasible if they can be trained on relatively simple datasets and still perform well on large, complex grids. As an example of a large and complex grid, we use a Texan power grid model and evaluate the models previously trained on dataset20 and dataset100. The benchmark models achieve surprisingly high performance with $R^2$ values above 84 \%; see the columns tr20evTexas and tr100evTexas in \Cref{tb_results_snbs}. Hence, the approach of training models on grids, which are smaller by more than one order of magnitude, is feasible. We want to emphasize that one successful attempt of a real-sized power grid should illustrate the general potential of this approach, but we still consider the hard evidence to be the generalization from 20 to 100 nodes. The performance is significantly better for the models trained on dataset100. We hypothesize that the repetition of geometrical structures more prevalent in dataset100 is useful for even larger grids. Grids of size 20 might still be too small to generalize to large grids, but the size 100 might actually be sufficient for many applications.\looseness-1

\subsection{Identification of troublemakers} \label{sec:id_tm}
In this section, we introduce a further benchmark tasks, namely to classify nodes into two categories, stable nodes or troublemakers as defined in \Cref{sec_method_trouble_makers}. As noted in \Cref{eq:TM max freq dev}, this target is essentially a thresholded version of the maximum frequency deviation (MFD). Thus, we can either directly train the classification task, or we can regress the MFD estimator and then threshold. Both strategies work and depending on the model different approaches seem to be best, see \Cref{tb_Results_TM_main_section}. The overall performance is very high; therefore, the prediction of TM is feasible. The GNNs ArmaNet and TAGNet outperform the baselines, and particularly TAGNet achieves good performance. \Cref{sec_TM_performance} contains additional results.

\begin{table*}[htb]
    \small
	\centering
	\caption{F2-score in \% for TM prediction. The column \emph{type} shows if classification (C) or regression (R) is used for training. For regression, thresholding is applied to compute the F2-score.}

	\begin{tabularx}{\linewidth}{Xp{.5cm}XXXXX}
		\toprule
	   Model & Type &tr20ev20 & tr100ev100 & tr20ev100 & tr20evTexas & tr100evTexas \\
		ArmaNet& C & \textbf{87.09} \tiny{$\pm$ 0.31} & 95.70 \tiny{$\pm$ 0.13} & 83.21 \tiny{$\pm$ 3.60} & 85.79 \tiny{$\pm$ 1.16} & 87.41 \tiny{$\pm$ 2.00} \\
        TAGNet & C & 86.07 \tiny{$\pm$ 0.25} & \textbf{96.62} \tiny{$\pm$ 0.01} & \textbf{96.53} \tiny{$\pm$ 0.03} & \textbf{93.07} \tiny{$\pm$ 0.39} & 91.84 \tiny{$\pm$ 0.29} \\
        ArmaNet & R&  83.12 \tiny{$\pm$ 0.07}  &  94.07 \tiny{$\pm$ 0.35}  &  92.16 \tiny{$\pm$ 0.63}  &  90.80 \tiny{$\pm$ 0.46}  &  94.79 \tiny{$\pm$ 0.99}\\
        TAGNet &  R& 85.69 \tiny{$\pm$ 0.23}  &  94.82 \tiny{$\pm$ 0.16}  & 92.85  \tiny{$\pm$ 0.15}  &  91.29 \tiny{$\pm$ 0.58}  &  \textbf{96.65} \tiny{$\pm$ 0.12} \\
        \midrule
        linreg & R & 72.73 & 91.51 & 91.12 & 73.22 & 93.75 \\
        MLP1 &  R& 74.41 \tiny{$\pm$ 0.01}  &  91.62 \tiny{$\pm$ 0.01}  &  51.24 \tiny{$\pm$ 0.48}  &  90.74 \tiny{$\pm$ 0.35}  &  92.71 \tiny{$\pm$ 0.42} \\
        MLP2 &  R& 74.38 \tiny{$\pm$ 0.04}  &  91.63 \tiny{$\pm 0.01 $}& 50.80  \tiny{$\pm$0.01 }  &  89.47 \tiny{$\pm$ 0.54}  & 93.51  \tiny{$\pm 0.12$} \\
	 \bottomrule
 	\end{tabularx}
 	\label{tb_Results_TM_main_section}
\end{table*}

\subsection{Benefits of using GNNs to predict dynamic stability}
As the experiments above show, GNNs are suitable for the analysis of the dynamic stability, both in terms of computational effort and predictive power. The performance exceeded our expectations. For SNBS and the best model (TAGNet), roughly 95 \% of the estimations differ only by 0.1 from the target values for tr100ev100 and tr100evTexas. This corresponds to the banded region shown in \Cref{fig_result_scatter_dataset_comparison}. The threshold of 0.1 can be motivated by considering the distributions in \Cref{fig_gridExamples_distributions}, because the modes are separated by 0.1. To achieve similar accuracy by conducting dynamical simulations with Monte-Carlo sampling, 100 perturbations per node are needed, which is a significant reduction to 10 000 perturbations we used for the generation of the datasets. With 100 perturbations, the computation for one grid of size 100 takes roughly 30 min and for the Texan grid 530 days using one CPU. Evaluating the GNNs takes less than 1\,s per grid. Hence, using GNNs is more than 1 800 times faster for grids of size 100 and 4.6 $\times 10^{7}$ times faster for the synthetic Texan power grid. This demonstrates the potential of analyzing many different configurations using GNNs. Furthermore, we show the successful prediction of TM, which can motivate future applications. \looseness-1

\section{Conclusion and Outlook}
In this work, we analyze the probabilistic dynamic stability of synthetic power grids using GNNs. We generate new datasets that are ten times larger than previous ones to enable the training of high capacity GNNs. Our benchmark results significantly improve over previous work and show that highly nonlinear SNBS can be predicted at surprisingly high accuracy by using more complex models and training on our larger dataset. We show that the models trained on our datasets can be used for prediction on a real-world-sized Texan power grid model. The results indicate the potential benefits of machine learning in this important domain. To ensure safe grid operation, dynamic assessments are necessary, but conventional methods require simulations that are too computationally expensive. GNNs can become helpful at analyzing almost unlimited many configurations due to their short evaluation times, once they are properly trained. This may turn out especially useful for grid expansion planning to quickly identify promising locations for new lines or generators. We expect the datasets to attract attention by research groups working on complexity science, non-linear dynamics, as well as machine-learning groups with a focus on GNNs. Furthermore, we introduce a new method to identify troublemakers in power grids and show their successful prediction. Besides further improving the performance, future work should try to explain the decision process of GNNs to generate new insights on the relation of topology and stability. Encouraged by our results, we will continue to extend our datasets with increasingly more complex and realistic grids, aiming at real power grids. For real-world applications, GNNs will have to prove that they can also be applied to realistic power grids using more complex node and line models. All data and code are publicly available  (see \Cref{appendix_source_code}). Given sufficient computational resources, the code can easily be adapted to generate more training data, or to simulate grids of different sizes. The open access enables the community to develop new methods to analyze future renewable power grids. \looseness-1

\section*{Acknowledgements}
All authors gratefully acknowledge the European Regional Development Fund (ERDF), the German Federal Ministry of Education and Research, and the Land Brandenburg for supporting this project by providing resources on the high-performance computer system at the Potsdam Institute for Climate Impact Research. Michael Lindner greatly acknowledges support by the Berlin International Graduate School in Model and Simulation (BIMoS) and by his doctoral supervisor Eckehard Schöll. Christian Nauck would like to thank the German Federal Environmental Foundation (DBU) for funding his PhD scholarship and Professor Raisch from Technical University Berlin for supervising his PhD. Special thanks go to Julian Stürmer for his assistance with the Texan power grid model.\looseness-1

\appendix
\section*{Appendix}

This section includes additional information to generate the datasets, reproduce the presented results and additional results that are not already shown in the main section. 

We start by providing information on the availability of the datasets and the used software, before providing more details regarding the modeling for the dataset generations. Subsequently, details regarding the training, the evaluation as well as the hyperparameter study are provided. Afterward, additional results are shown.

\subsection{Availability of the datasets}
\label{appendix_source_code}
The code to generate the datasets and train ML models is available at GitHub \url{https://github.com/PIK-ICoNe/dynamic_stability_datasets_gnn_paper-companion}. The datasets prepared for the training of ML applications can be found at Zenodo \url{10.5281/zenodo.8204334}. The data is publicly available and licensed under the Creative Commons Attribution 4.0 International license (CC-BY 4.0). The code to generate figures, including the trained models, is available at: Zenodo: \url{10.5281/zenodo.8205284}.

\subsection{Software for generating the datasets}
Julia is used for the simulations \cite{bezansonJuliaFreshApproach2017} and the dynamic simulations rely on the package DifferentialEquations.jl \cite{rackauckasDifferentialEquationsJlPerformant2017}. For simulating more realistic power grids in future work, we recommend the additional use of NetworkDynamics.jl \cite{lindnerNetworkDynamicsJlComposing2021} and PowerDynamics.jl \cite{plietzschPowerDynamicsJlExperimentally2021}.

\subsection{Software for training}
The training is implemented in PyTorch \citep{paszkePyTorchImperativeStyle2019}. For the graph handling and graph convolutional layers, we rely on the additional library PyTorch Geometric \citep{feyFASTGRAPHREPRESENTATION2019}. The models are fit with the SGD-optimizer. As loss function, we use the mean squared error(MSELoss in PyTorch) for regression, and binary cross entropy with an included Sigmoid layer (BCEWithLogitsLoss) for classification. Furthermore, \texttt{ray} \citep{moritzRayDistributedFramework2018} is used for parallelizing the hyperparameter study.

\subsection{Modeling details of generating the datasets}
\label{sec_modelling_details_dataset_generation}

In the detail of the modeling and the size of the dataset, we attempt to strike a balance between relevance to real-world applications, computational tractability and conceptual simplicity. Therefore, we employ the following criteria: (i) generate synthetic network topologies that mimic real-world power grids; (ii) model the main dynamics of self-organized power flow and synchronization; (iii) minimize statistical and numerical errors with highly accurate simulations; (iv) to study out-of distribution tasks and scale effects, consider grid sizes of different orders of magnitude. \looseness-1

The most important simplifications in comparison with real-world power grids are homogeneous edges, fixed magnitudes of sources/sinks and modeling all nodes by the swing equation. In contrast to our modeling, real power grid lines have different properties and there are more complex models for generators and loads. However, previous studies have shown that many interesting observations are still possible under our assumptions \cite{nitzbonDecipheringImprintTopology2017}. 

To investigate different topological properties of differently sized grids, we generate two datasets with either 20 or 100 nodes per grid, referred to as dataset20 and dataset100. To enable the training of complex models, both datasets consist of 10,000 graphs. Additionally, probabilistic dynamic stability values of a synthetic model of the Texan power grids are provided for evaluation purposes.

\paragraph{Modeling of the power grids}
This section covers the precise modeling of power grids used for the dataset generation and may be skipped. To generate realistic topologies, we use the package \texttt{SyntheticNetworks}\citep{schultzRandomGrowthModel2014,schultzHttpsGithubCom2020}. The sources and sinks are assigned randomly. For the dynamic simulations, all nodes are represented by the second-order-Kuramoto model \citep{kuramotoSelfentrainmentPopulationCoupled2005,rodriguesKuramotoModelComplex2016}, which is also called a swing equation, see \Cref{eqKuramoto}.
Using homogeneous coupling strength ($K$) can be interpreted as considering power grids that only have one type of power line and comparable distances between all nodes. This assumption is justifiable, because in real power grids longer lines are built stronger; therefore the actual coupling does not scale as much with the length.

To estimate SNBS we rely on the approach presented in \cite{nauckPredictingBasinStability2022}:
\begin{displayquote}
"[F]or every node in a graph, $M = 10,000$ samples of perturbations per node are constructed by sampling a phase and frequency deviation from a uniform distribution with $(\phi, \dot{\phi}) \in [- \pi, \pi] \times [-15, 15]$ and adding them to the synchronized state. Each such single-node perturbation serves as an initial condition of a dynamic simulation of our power grid model, [cf. \Cref{eqKuramoto}]. At $t=500$ the integration is terminated and the outcome of the Bernoulli trial is derived from the final state. A simulation outcome is referred to as \emph{stable} if at all nodes $\dot{\phi_i} < 0.1$. Otherwise, it is referred to as \emph{unstable}. The classification threshold of $0.1$ is chosen, accounting for minor deviations due to numerical noise and slow convergence rates within a finite time-horizon."
\end{displayquote}

To ensure the reliability of our results, we try to minimize numerical and statistical errors: For the dynamical simulations, a higher order Runge-Kutta methods with adaptive time stepping, and low error tolerances is used. For the Monte-Carlo sampling, 10 000 simulations per node result in standard errors for our probabilistic measure SNBS of less than $\pm 0.01$.

Furthermore, we want to provide some numbers regarding the simulation time. The computation of a single perturbation in case of dataset20 takes 0.056\,s , in case of dataset100 0.189\,seconds and in case of the Texan power grid 239.97\,s.

\subsection{Error bounds for TM prediction}
\label{app_TM_error_bounds}
For estimating the troublemaker property, we use the same simulations as for the SNBS estimation. Due to the smaller region of initial conditions, this reduces the minimum number of available simulations per node to 1 595 for dataset20, 1 569 for dataset100 and 135 for the Texan power grid. For 5 nodes in the Texan grid, less than 135 of the previously simulated perturbations had initial magnitude $< 2.5$. To reach the minimum amount of 135 samples per node, we simulated additional trajectories for these nodes (16 in total). Taking this data limitation into account, we choose $\gamma = 0.005$ for dataset20 and dataset100 and $\gamma = 0.05$ for the Texan grid.

\subsection{Evaluation of the performance using different metrics}
\label{sec_evaluation_metrics}

Define the \emph{mean squared error ($\mathrm{mse}$)} between $n$-dimensional predicted values $f$ and target values $y$ as 
\begin{equation}
	\mathrm{mse}(f,y) := \frac{1}{n} \sum_{i=1}^n (f_i - y_i)^2
\end{equation}

As an evaluation metric for the regression, we use the \emph{coefficient of determination $R^2$}
\begin{equation}
	R^2 := 1 - \frac{\mathrm{mse}(f,y)}{\mathrm{mse}(y_{mean},y)},
\end{equation}
where $y_{mean}$ is the mean of the target values of the test dataset. The $R^2$-score captures the mean squared error relative to a null model that predicts the mean of the test dataset for all points. Thus, the $R^2$-score measures the percentage of variance of the data explained by the prediction. By design, a model that predicts the mean of the target values has $R^2 = 0$. 

For classification, we use the $F_2$-score, which is based on recall and precision. The recall of a classifier is defined as $\frac{TP}{TP+FN}$, where $TP$ denotes true positives and $FN$ false negatives, and the precision is defined as $\mathrm{precision} = \frac{TP}{TP+FP}$. Finally, the $F_\beta$-score is
\begin{equation}
F_\beta := (1 + \beta^2) \cdot \frac{\mathrm{precision} \cdot \mathrm{recall}}{\beta^2 \cdot \mathrm{precision} + \mathrm{recall}}.
\end{equation}

The $F_2$-score gives more weight to recall and less to the precision. This is an appropriate metric for  identifying vulnerabilities of power grids since it is more important to identify all critical states, even if this increases the number of false positives.

\subsection{Hyperparameter optimization}
\label{sec_hyperparameter}
We conduct hyperparameter studies in two steps. First, we optimize model properties such as the number of layers and channels as well as layer-specific parameters e.g. the number of stacks and internal layers in the case of ArmaNets. For this optimization, we use dataset20 and the SNBS task only. For all models we investigated the influence of different numbers of layers and the numbers of channel between multiple layers. We limit the model size to just above $4\times 10^6$ parameters; therefore, we did not investigate the full presented space, but limited, for example, the number of channels when adding more layers. The resulting models have the following properties: ArmaNet has 3 layers and 189,048 parameters. GCNNet has 7 layers and 523,020 parameters. SAGENet has 8 layers and 728,869 parameters. TAGNet has 13 layers and 415,320 parameters.

Afterward, we optimize the learning rate, batch size, and scheduler of the best models for dataset20 and dataset100 and the tasks SNBS/TM separately. Hence, our models are not optimized to perform well at the out-of distribution task. The best model from \citep{nauckPredictingBasinStability2022} referred to as Ar-bench is used as another baseline. It is a GNN model consisting of 1 050 parameters and based on two Arma-layers. The only adjustment to that model is the removal of the fully connected layer after the second Arma-Convolution and before applying the Sigmoid layer, which improves the training.

\subsubsection{Baselines: MLP}
\label{appendix_MLP}
MLP1 has one hidden layer with 35 units per layer, resulting in 1 541 parameters, and MLP2 has six hidden layers with 500 units per layer, leading to 1 507 001 parameters. We conducted hyperparameter studies to optimize the batch sizes and learning rates. To scale the input features, node-wise standardization among grids with the same number of nodes is used. The nodes in dataset20 and dataset100 are standardized with the mean and standard deviation in the respective training sets. For the Texan grids, all available nodes are considered.

\subsection{Details of the training of the benchmark models}
To reproduce the obtained results, more information regarding training is provided in this section. Detailed information on the training as well as the computation time is shown in \Cref{tb_computation_time_dataset20_dataset100}. In the case of dataset20, a scheduler is not applied; in the case of dataset100, schedulers are used for Ar-bench (stepLR), GCNNet (ReduceLROnPlateau). The default validation and test set batch size is 150. The validation and test batchsize for Ar-bench and ArmaNet is 500 in the case of dataset20 and 100 for dataset100. The number of trained epochs differs, because the training is terminated in the case of significant overfitting. Furthermore, different batch sizes have a significant impact on the simulation time. Most of the training success occurs within the first 100 epochs; afterward, the improvements are relatively small.

\begin{table*}[htb]
	\centering
	\caption{Properties of training models on a SNBS task. Regarding the training time: we train five seeds in parallel using one nVidia V100.}
    \begin{tabularx}{\linewidth}{XXXXXXXXX}
		\toprule
    name & \multicolumn{2}{l}{number of epochs} & \multicolumn{2}{l}{training time} & \multicolumn{2}{l}{train batch size} & \multicolumn{2}{l}{learning rate} \\
		dataset& 20 & 100 & 20 (hours) & 100 (days)  & 20 & 100 & 20 & 100\\
		\midrule
		Ar-bench & 1,000 & 800 & 26 & 4 & 200 & 12 & 0.914 & .300 \\ 
		ArmaNet & 1,500 & 1,000 & 46 & 6 & 228 & 27 & 3.00 & 3.00 \\
		GCNNet & 1,000 & 1000& 29 & 5 & 19 &79 & .307 &.286 \\
		SAGENet & 300 & 1000 & 9 & 5 & 19 & 16 & 1.10 &  1.23 \\
		TAGNet & 400 & 800 & 11 & 4 &  52 & 52 & 0.193 &  .483 \\
 	\end{tabularx}
	\label{tb_computation_time_dataset20_dataset100}
\end{table*}

\subsection{Visualization of the ArmaNet at predicting SNBS}
\begin{figure*}
    \centering
        \includegraphics[width=\linewidth]{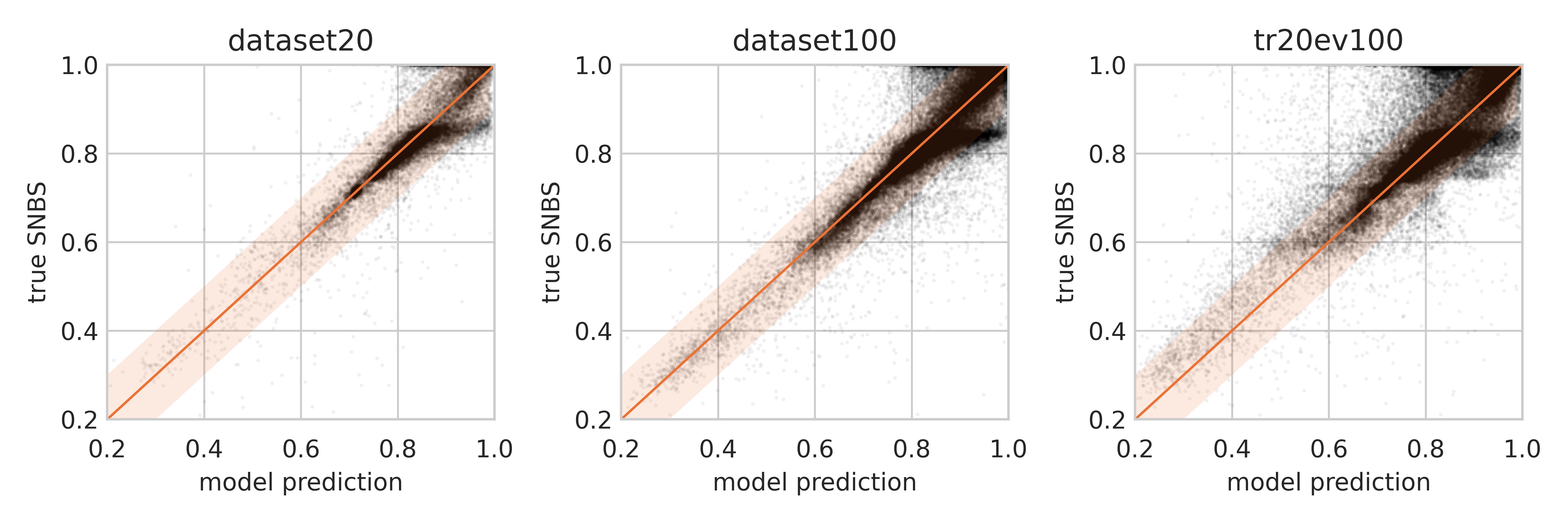}
    \vspace{-10pt}
    \caption{SNBS over a predicted output of the ArmaNet model for the in-distribution tasks on dataset20 and dataset100 and the out-of-distribution generalization from dataset20 to dataset100 (tr20ev100). The diagonal represents a perfect model ($R^2 = 1$), and the banded region indicates predictions, which are accurate up to an error of $\pm0.1$.}
    \vspace{-11pt}
	\label{fig_result_scatter_Arma}
\end{figure*}

\Cref{fig_result_scatter_Arma} visualizes the performance of ArmaNet3 at the in-distribution tasks tr20ev20, tr100ev100 as well as the out-of-distribution generalization tr20ev100.

\subsection{Detailed results of training on a smaller dataset}
\label{sec_appendix_smaller_dataset}
To investigate the influence of available training data and to connect with the previous work, we train all models on only 800 grids, from \cite{nauckPredictingBasinStability2022}. The results are shown in \Cref{tb_ResultsR2score_small_data}. 

\begin{table*}[htb]
	\centering
	\caption{Performance after training on a smaller training set. All models are trained on the same 800 grids as in \cite{nauckPredictingBasinStability2022}, but evaluated on a newly introduced test set. The results are represented by an $R^2$ score in \%.}
	\begin{tabularx}{\linewidth}{XXXX}
		\toprule
	   Model & tr20ev20 & tr100ev100 & tr20ev100\\
		\hline
		Ar-bench & 46.74 \tiny{$\pm$ 2.31} & 56.57 \tiny{$\pm$ 1.89} & 33.78 \tiny{$\pm$ 1.86} \\
		ArmaNet & \textbf{73.45} \tiny{$\pm$ 0.22}& \textbf{83.65} \tiny{$\pm$ 0.48}& 57.51  \tiny{$\pm$ 0.50}   \\
        GCNNet & 49.61 \tiny{$\pm$ 0.43}  & 63.59 \tiny{$\pm$ 0.28}& 38.53 \tiny{$\pm$ 0.77}   \\
        SAGENet & 49.51 \tiny{$\pm$ 0.50} & 60.64 \tiny{$\pm$ 0.06} & 38.37  \tiny{$\pm$ 0.25} \\
        TAGNet   & 73.18 \tiny{$\pm$ 0.30} &  82.35 \tiny{$\pm$ 0.13}  &  \textbf{60.60} \tiny{$\pm$ 2.18}\\
	 \bottomrule
 	\end{tabularx}
 	\label{tb_ResultsR2score_small_data}
\end{table*}

\subsection{Performance of identifying troublemakers}
\label{sec_TM_performance}
Last, we show the additional results and training details of predicting the troublemakers using classification and regression setup. We use the previously introduced GNN models (see \Cref{sec_Experimental_setup}). For each regression and classification task, we conduct another hyperparameter study to optimize the learning rates. The same inputs are used: adjacency matrix $A$, and nodal power input $P$. For identifying troublemakers, a thresholded variant of the semi-analytic lower bound for survivability \citep{hellmannSurvivabilityDeterministicDynamical2016} is used as a further baseline. This bound can be directly computed from the input data and, hence, requires no training.

Besides introducing two classes for predicting troublemakers, it is also possible to directly predict the maximum frequency deviation per node (\Cref{fig_gridExamples_distributions}). The results are shown in \Cref{tb_results_TM_R2}. The predictions of the regression can be complemented by applying thresholding afterward to categorize nodes as troublemakers. After applying the thresholding, the results can again be evaluated using the F2-score (\Cref{tb_Results_TM_main_section}). Further baselines are shown in \Cref{tb_Results_TM_appendix}. We use weighted loss functions to account for the small number of TM-nodes. 

\begin{table*}[htb]
    \small
	\centering
	\caption{F2-score in \% for TM prediction. The column \emph{type} shows if classification (C) or regression (R) is used for training. For regression, thresholding is applied to compute the F2-score. See \Cref{tb_Results_TM_main_section} for more models.
 (*) Since the semi-analytic baseline does not require training, its performance is directly evaluated on the test set.}
	\begin{tabularx}{\linewidth}{Xp{.5cm}XXXXX}
		\toprule
	   Model & Type &tr20ev20 & tr100ev100 & tr20ev100 & tr20evTexas & tr100evTexas \\
        logreg & C & 44.13 & 73.69 & 54.43 & 18.52 & 54.67\\
        MLP1 & C & 74.58 \tiny{$\pm$ 0.17} & 91.64 \tiny{$\pm$ 0.01} & 52.59 \tiny{$\pm$ 0.52} & 90.96 \tiny{$\pm$ 1.02} & 87.07 \tiny{$\pm$ 0.64} \\
        MLP2 & C & 74.46 \tiny{$\pm$ 0.03} & 91.61 \tiny{$\pm$ 0.00} & 55.22 \tiny{$\pm$ 1.27} & 90.67 \tiny{$\pm$ 0.39} & 85.40 \tiny{$\pm$ 0.19} \\
        Semi-analytic & - & 22.86* & 38.53* & 38.53*& 15.73* & 15.73* \\
	 \bottomrule
 	\end{tabularx}
 	\label{tb_Results_TM_appendix}
\end{table*}

\begin{table*}[htb]
    \small
	\centering
	\caption{Results of predicting maximum frequency deviations represented by $R^2$ score in \%.}
	\begin{tabularx}{\linewidth}{XXXXXX}
		\toprule
	   model & tr20ev20 & tr100ev100 & tr20ev100 & tr20evTexas & tr100evTexas \\
		ArmaNet &  \textbf{96.48} \tiny{$\pm$ 0.05}  &  97.60\tiny{$\pm$ 0.02}  &  \textbf{95.36} \tiny{$\pm$ 0.23 }  &  \textbf{84.36} \tiny{$\pm$ 1.67}  &  89.41 \tiny{$\pm$ 1.01} \\
        TAGNet & 96.40 \tiny{$\pm$ 0.06}  & \textbf{97.88}  \tiny{$\pm$ 0.04}  &  93.60\tiny{$\pm$ 0.12}  & 83.75 \tiny{$\pm$ 0.48}  &  \textbf{95.24} \tiny{$\pm$ 0.65} \\
        \midrule
        linreg & 83.87 & 87.77 & 86.61 & 80.80 & 78.11 \\
        MLP1 &  90.87 \tiny{$\pm$ 0.02}  &  93.58 \tiny{$\pm$ 0.02}  &  86.38 \tiny{$\pm$ 0.27}  &  67.90 \tiny{$\pm$ 7.11}  &  65.20 \tiny{$\pm$ 6.97} \\
        MLP2 &  90.75 \tiny{$\pm$ 0.02}  &  93.61  \tiny{$\pm$ 0.00} &  87.37 \tiny{$\pm$ 0.23}  &  82.96 \tiny{$\pm$ 0.19}  &  86.63 \tiny{$\pm$ 0.08}\\
	 \bottomrule
 	\end{tabularx}
 	\label{tb_results_TM_R2}
\end{table*}

In addition to an F2-Score shown in \Cref{tb_Results_TM_main_section}, we provide the recall as a further performance indicator in \Cref{tb_Results_TM_recall_appendix}. The GNNs outperform the baselines in all tasks, and particularly, TAGNet achieves a good performance. GNNs achieve high recalls throughout all tasks, including the out-of-distribution generalizations, while still keeping high F2-scores. 

\begin{table*}[htb]
    \small
	\centering
	\caption{Recall in \% for TM prediction. The column \emph{type} shows if classification (C) or regression (R) is used for training. For regression, thresholding is applied to compute the F2-score.}
	\begin{tabularx}{\linewidth}{Xp{.5cm}XXXXX}
		\toprule
	   Model & Type &tr20ev20 & tr100ev100 & tr20ev100 & tr20evTexas & tr100evTexas \\
		ArmaNet & C & 89.52 \tiny{$\pm$ 0.47}  &  96.54 \tiny{$\pm$ 0.20} &  85.08 \tiny{$\pm$ 4.84} &  95.42 \tiny{$\pm$ 2.36}  & 97.39 \tiny{$\pm$ 0.65}\\
        TAGNet & C & 86.06 \tiny{$\pm$ 0.37} &  97.59 \tiny{$\pm$ 0.06} & 97.31 \tiny{$\pm$ 0.13} & \textbf{100.00} \tiny{$\pm$ 0.00} & \textbf{100.00} \tiny{$\pm$ 0.00} \\
        ArmaNet & R & 80.57 \tiny{$\pm$ 0.07}  &  93.14 \tiny{$\pm$0.42}  & 92.52 \tiny{$\pm$ 0.87}  &  94.12 \tiny{$\pm$ 0.00}  & 97.39 \tiny{$\pm$ 0.65} \\
        TAGNet & R & 83.59  \tiny{$\pm$ 0.26}  & 94.03  \tiny{$\pm$ 0.24}  &  94.83 \tiny{$\pm$ 0.24}  &  97.39 \tiny{$\pm$ 1.31}  & 98.04  \tiny{$\pm$ 0.65} \\
        \midrule
        logreg & C & 86.93 & 91.82 & 95.55 & 98.03 & 94.12\\
        MLP1 &  C & 70.32 \tiny{$\pm$ 0.25} & 89.91 \tiny{$\pm$ 0.01} & 47.08 \tiny{$\pm$ 0.53} &  93.46 \tiny{$\pm$ 1.73} &  94.12 \tiny{$\pm$ 0.00} \\
        MLP2 &  C & 70.04 \tiny{$\pm$ 0.04} & 89.87  \tiny{$\pm$ 0.01} & 49.70 \tiny{$\pm$ 1.29} &  91.50 \tiny{$\pm$ 0.65} &  94.77 \tiny{$\pm$ 0.65} \\
        linreg & R& 70.55 & 89.75 & 89.29 & 68.63 &  94.12\\
        MLP1 &  R & 69.96 \tiny{$\pm$ 0.00}  &  89.89 \tiny{$\pm$ 0.01}  & 45.74 \tiny{$\pm$ 0.54}  &  92.16 \tiny{$\pm$0.00}  &  94.77 \tiny{$\pm$ 0.65} \\
        MLP2 & R & 69.93  \tiny{$\pm$ 0.04}  &  89.90 \tiny{$\pm$ 0.01} &  45.24 \tiny{$\pm$ 0.01}  &  88.89 \tiny{$\pm$ 0.65}  & 94.12  \tiny{$\pm$0.00}\\
        semi-analytic & & \textbf{92.70}* & \textbf{97.77}* & \textbf{98.04}* & 97.77* & 98.04* \\
	 \bottomrule
 	\end{tabularx}
 	\label{tb_Results_TM_recall_appendix}
\end{table*}

\FloatBarrier
%

\end{document}